\documentclass[review]{elsart}
\usepackage{natbib}
\usepackage{graphicx}
\usepackage{endfloat}
\usepackage{subfigure}
\usepackage{amssymb}
\usepackage{lineno}
\begin{document}
\begin{frontmatter}

\title{Network-Scale Traffic Modeling and Forecasting with Graphical Lasso and Neural Networks}
\author{Shiliang Sun,}
\author{Rongqing Huang\corauthref{cor1}}
\corauth[cor1]{Corresponding author. Tel.: +86-21-54345186; fax:
+86-21-54345119. \\{\it E-mail address:} {\rm rqhuang09@gmail.com} (R. Huang).}
\author{, and Ya Gao}
\address{Department of Computer Science and Technology, East China Normal University, 500 Dongchuan Road, Shanghai 200241, China}

\begin{abstract}
Traffic flow forecasting, especially the short-term case, is an important topic in intelligent transportation systems (ITS).
This paper does a lot of research on network-scale modeling and forecasting of short-term traffic flows. Firstly, we propose
the concepts of single-link and multi-link models of traffic flow forecasting. Secondly, we construct four prediction models
by combining the two models with single-task learning and multi-task learning. The combination of the multi-link model and
multi-task learning not only improves the experimental efficiency but also the prediction accuracy. Moreover, a new multi-link
single-task approach that combines graphical lasso (GL) with neural network (NN) is proposed. GL provides a general methodology
for solving problems involving lots of variables. Using L1 regularization, GL builds a sparse graphical model making use of the
sparse inverse covariance matrix. In addition, Gaussian process regression (GPR) is a classic regression algorithm in Bayesian
machine learning. Although there is wide research on GPR, there are few applications of GPR in traffic flow forecasting. In this paper,
we apply GPR to traffic flow forecasting and show its potential. Through sufficient experiments, we compare all of the proposed approaches
and make an overall assessment at last.
\end{abstract}
\end{frontmatter}

\section*{Introduction}
With the accelerated pace of modern life, more and more cars come into use. The increased use of cars brings convenience to the public on
the one hand, but on the other hand produces many social problems such as traffic congestion, traffic accidents and environmental
pollution. To make traffic management and traveler information services more efficient, intelligent transportation systems (ITS) emerges as
the times required, and the manual traffic management is no longer viable. The benefits of ITS can not be realized without the ability to
anticipate short-term traffic conditions. As an important aspect of traffic conditions, traffic flows can give insights into traffic conditions\citep{YuMarkov03}.
Therefore, short-term traffic flow forecasting becomes one of the most important and fundamental problems in ITS. Short-term traffic flow forecasting is to
determine the traffic flows in the next time interval, usually in the range of five minutes to half an hour, using historical data\citep{AbdulhaiNN02,Sun07}.
A good short-term traffic flow forecasting model can tell the right traffic condition in the near future and make traffic management more effective in turn.
In recent years, ITS, especially short-term traffic flow forecasting has already attracted great interest of researchers. Our work also focuses on the topic
of short-term traffic flow forecasting.

The items detected by ITS generally include traffic flow, volume and occupancy etc. Among all these items, traffic flow is considered to
be the typical metric of traffic condition on a certain link~\citep{ChenEnsemble07}. Traffic flow measures the number of vehicles passed
through in a defined time interval and lower traffic flow means heavier traffic congestion. Traditional traffic flow forecasting predicts
a future flow of a certain link only using the historical data on the same link, which is also called single-link traffic flow forecasting.
Obviously, single-link forecasting approaches ignore the relations between the measured link and its adjacent links. In fact, each link is
closely related to other links in the whole transportation system, especially their adjacent links. In this paper, we put forward the multi-link
forecasting models which take the relations between adjacent links into account. Sufficient experiments on real world data show that multi-link
approaches are superior to single-link approaches.

In the past decade, series of traffic flow forecasting approaches have been proposed,
such as time series based approaches~\citep{MoorthyShort88,LeeSubsets99,WilliamSeasonal03},
nonparametric methods~\citep{DavisRegression91}, local regression models~\citep{Davis90,SmithTraffic97},
neural network approaches~\citep{HallArtificial98}, Kalman filtering~\citep{OkutaniDynamic84}, Markov chain model~\citep{YuMarkov03} and so on.
Among all these approaches, neural-network-based forecasting approaches are considered as relatively effective methods due to their well-established models.
Typical neural-network-based forecasting methods mainly include back propagation (BP) neural network~\citep{SmithTrafficApproaches94}, radial basis
function (RBF) neural network~\citep{WangRadialBasis03,ParkRadialBasis98}, recurrent neural network~\citep{UlbrichtMulti94}, time delayed neural network~\citep{AbdulhaiNN99}, resource allocated networks~\citep{ChenSequential01}, etc. In this paper, we select BP neural networks to serve as the corresponding neural-network-based experiments.
The competitive results further verify the superiority of the proposed neural-network-based approaches.

Gaussian process regression (GPR) is a classic regression
algorithm basing on Bayesian theory. A Gaussian process is a
generalized Gaussian probability distribution and each process is
specified by its mean function and covariance function~\citep{RasmussenGaussian06}. Because of
the characteristics of easy implementation, few parameters and
strong interpretability, GPR is studied widely in machine learning.
Furthermore, theoretical and practical developments over the last
decade have shown that Gaussian process is a serious competitor for
supervised learning applications~\citep{RasmussenGaussian06}. However, there are few
applications of GPR in traffic flow forecasting. In this paper, we give a brief analysis of GPR and apply it to
traffic flow forecasting. Through sufficient tests of GPR in real-world data sets, we
point out the potential of GPR for traffic flow forecasting.

Graphical model is not rare in both statistics and computer science.
It is considered as an intersection of the two fields. In
statistics applications, there are often large-scale models with
thousands or even millions of variables involved. Similarly,
there are the same problems in machine learning applications, such
as biological information retrieval, language processing and so on.
Graphical lasso (GL) provides a general methodology for solving such
problems~\citep{JordanGraphical04}. By using L1 regularization, GL
builds a sparse graphical model making use of the sparse inverse
covariance matrix. In this paper, we provide a detailed discussion of the GL algorithm in theory and apply it to multi-link traffic flow
forecasting. With the further information extracted by GL, and combining with the BP neural networks,
we construct a new multi-link single-task traffic flow prediction model, which we refer to as GL\_NN.

Parts of our work have been presented recently at international conferences~\citep{Gao2010,Gao2011}.
In this paper, we combine and extend them to give a more systematical analysis.
The remainder of this paper is organized as follows. First, we introduce the four prediction models
basing on NNs. Next, we give the introduction of GPR and GL, respectively, which are closely
related to our work. Then, all the corresponding experiments and discussions on GPR are
presented in the section of experiments.  Finally, conclusions are given in the last section.

\section*{Prediction Models with Neural Networks}
Due to the excellent ability in handling complex problems, and the
characteristics of self-learning, self-organizing and self-adaptation,
neural networks (NNs) usually perform well in machine learning problems. On the other hand,
multi-task learning (MTL) is widely applied in computational intelligence and has shown competitive performance.
The main difference between MTL and single-task learning (STL) can be that, with the same inputs, MTL has multiple outputs but STL
has only one output at a time. As to the multiple tasks in MTL, there is only one main task and the others are extra tasks assisting
the learning of the main task. More details about MTL and STL can be found in~\citet{Caruana97}.
In this paper, we further combine the single-link and multi-link models with single-task learning and multi-task learning to
construct four prediction models. The four models are single-link single-task learning (SSTL), single-link
multi-task learning (SMTL), multi-link single-task learning (MSTL)
and multi-link multi-task learning (MMTL), respectively.

\subsection*{Single-Link Model}
Traditional traffic flow prediction models are single-link models, which predict the future flow of one certain road link using only the historical data of the same link. Combining the single-link model with single-task learning and multi-task learning, we construct two models which are SSTL and SMTL,  respectively. The main difference of the two models lies in the different numbers of outputs, which is also the difference of STL and MTL approaches in a very narrow sense. Following the settings in~\citet{Caruana97}, we set the number of outputs as 3(one main task and
two extra tasks) for our MTL approaches. For one link, we use the first 5 historical traffic flows to predict the next one. That is, the number of inputs is 5.

\begin{figure}[thb]
\centering
\includegraphics[scale=0.5]{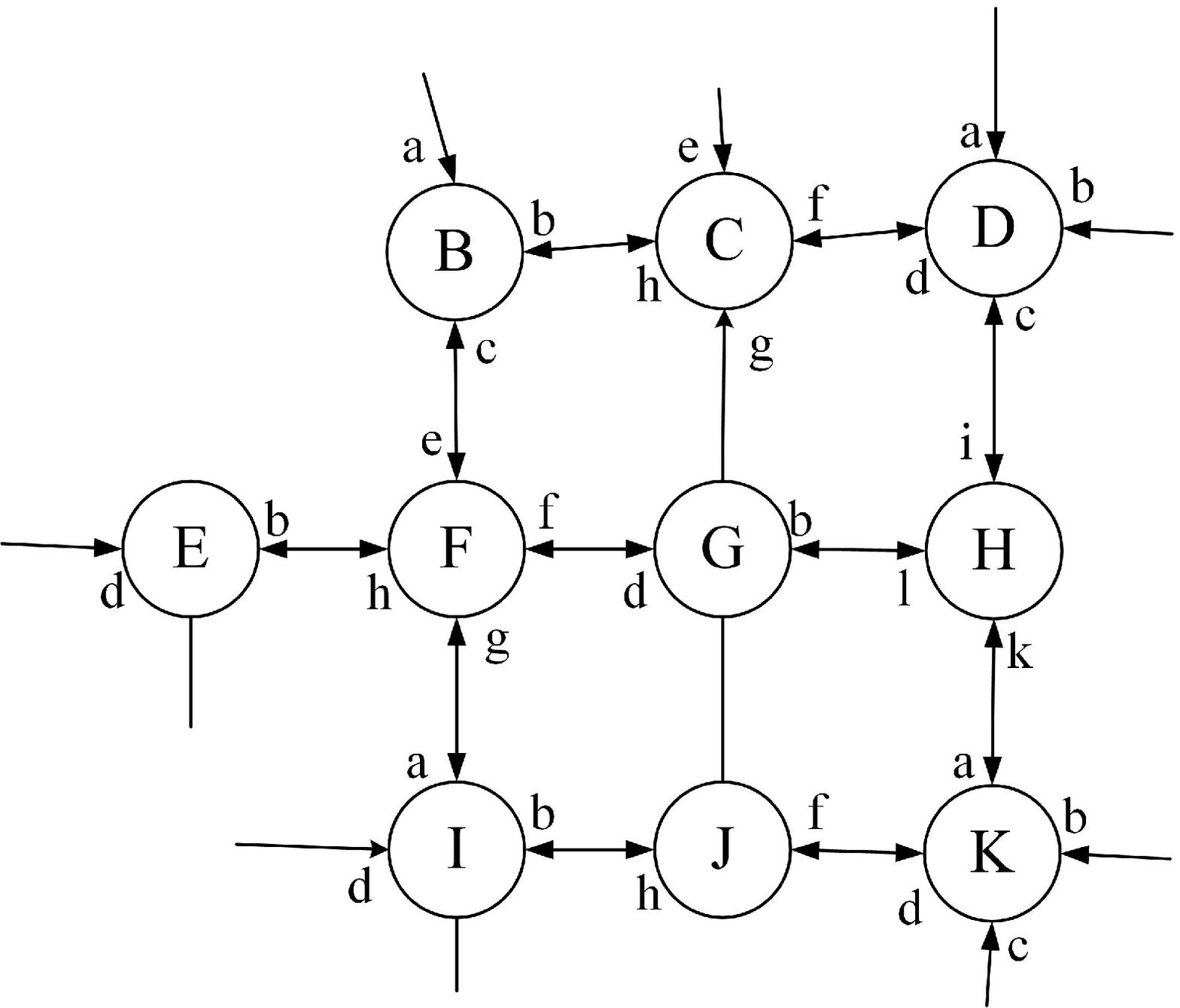}
 \caption{The sketch map of road links used in the paper.}
 \label{Figure 1}
\end{figure}
Take link Ba in Fig.~1 as an example. Record the traffic flow of road link Ba at time interval $n$ as $t_a(n)$, and then
the corresponding five historical traffic flows are respectively $t_a(n-5),\cdots, t_a(n-1)$. In the single-link model, we predict $t_a(n)$ using
$t_a(n-5), \cdots, t_a(n-1)$. Basing on NNs, the five historical flows are five inputs. In SSTL, $t_a(n)$ is the one and only output.
While in SMTL, there are three outputs $t_a(n-1), t_a(n)$ and $t_a(n+1)$. Among the three outputs, $t_a(n)$ is the main task,
$t_a(n-1)$ and $t_a(n+1)$ are extra tasks to assist the prediction of the main task. Note that the selection of extra tasks are not
specified. Here we follow the settings in previous experiments~\citep{Jin08}. Diagrams of the two single-link models SSTL and SMTL are
shown in Fig.~2a and Fig.~2b.

\begin{figure}[thb]
\centering
 \subfigure[Single-link single-task learning (SSTL) model]{
\includegraphics[scale=0.5]{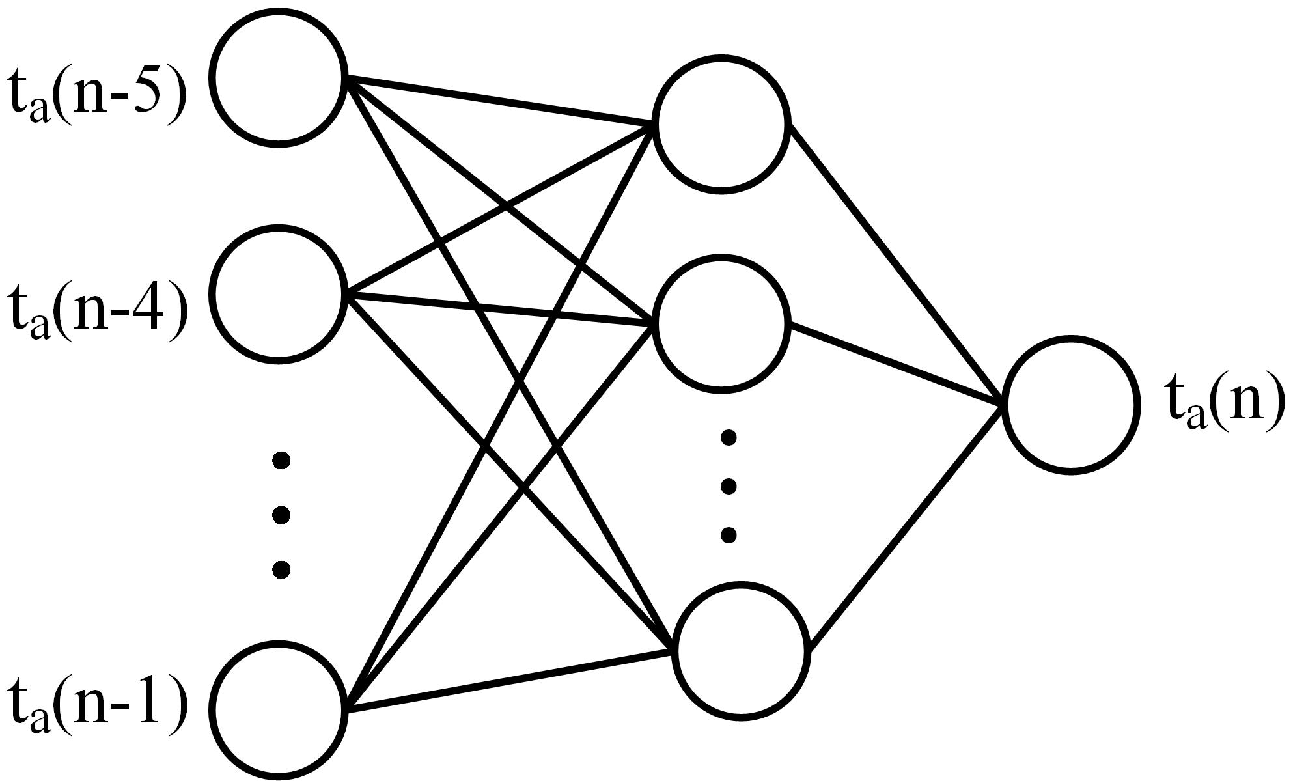}
} \subfigure[Single-link multi-task learning (SMTL) model]{
\includegraphics[scale=0.5]{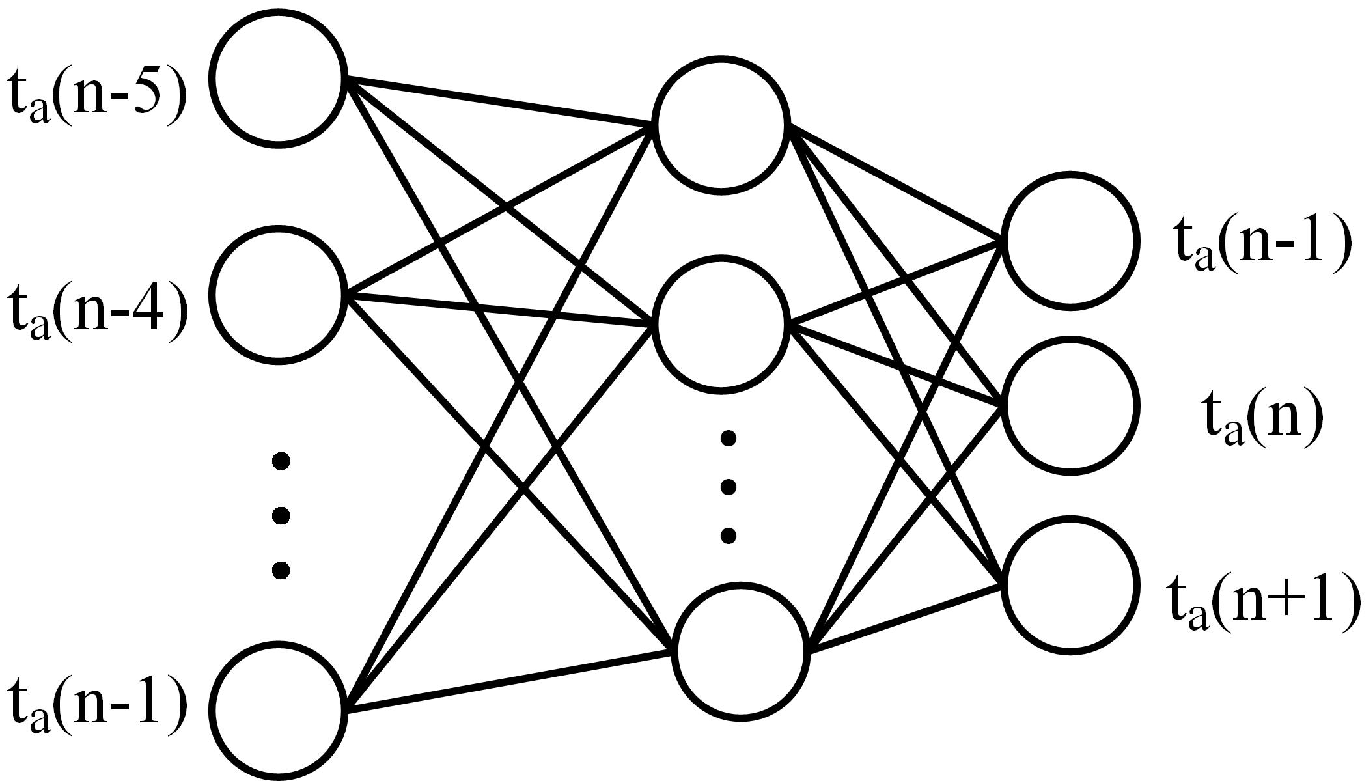}}
 \caption{Diagrams of the two single-link models based on NNs.}
 \label{Figure 2}
\end{figure}

\subsection*{Multi-Link Model}
Obviously, the single-link prediction model is inefficient because it just predicts one future flow of only one link at a time.
Worse still, it does not make use of the relevant information between adjacent links to improve the prediction.
In fact, since vehicles always come from one link and go to other links in the whole transportation system, traffic
flows of all links, especially adjacent links in the whole traffic network are relevant. Therefore, taking the relevance between adjacent links into account,
we combine the multi-link model with single-task learning and multi-task learning to construct the multi-link single-task learning model (MSTL)
and the multi-link multi-task learning model (MMTL). The multi-link model can simultaneously predict multiple traffic flows of multiple links using historical flows of all links
in the same junction at a time. Certainly, using the historical data from multiple links in different junctions to predict the flow on one link of them is the special case of the multi-link model. For example, GL\_NN is a special case of the multi-link model.

Again take the map in Fig.~1 as an example. We can see that the junction B connects three links Ba, Bb and Bc. In multi-link
models, we simultaneously predict future flows of the three links, using all the historical data of the three links. Therefore,
similar to the analysis in single-link model, there are 3$\times$5=15 inputs in multi-link models, which are respectively
$t_a(n-5), \cdots, t_a(n-1), t_b(n-5), \cdots, t_b(n-1)$, and $t_c(n-5), \cdots, t_c(n-1)$. Combing with single-task learning and multi-task learning, there
are three outputs $t_a(n), t_b(n)$ and $t_c(n)$ in MSTL, while 3$\times$3=9 outputs $t_a(n-1), t_a(n), t_a(n+1), t_b(n-1), t_b(n), t_b(n+1), t_c(n-1), t_c(n)$ and $ t_c(n+1)$ in
MMTL. Diagrams of the two multi-link models are shown in Fig.~3a and Fig.~3b. For the sake of clarity, in Fig.~3b, we draw the three outputs corresponding to the three links Ba,
Bb and Bc each in a box. Through the four diagrams shown as Fig.~2a, Fig.~2b, Fig.~3a and Fig.~3b, we can get a better understanding of single-link and multi-link models.

\begin{figure}[thb]
\centering
\subfigure[Multi-link single-task learning (MSTL) model]{
\includegraphics[scale=0.5]{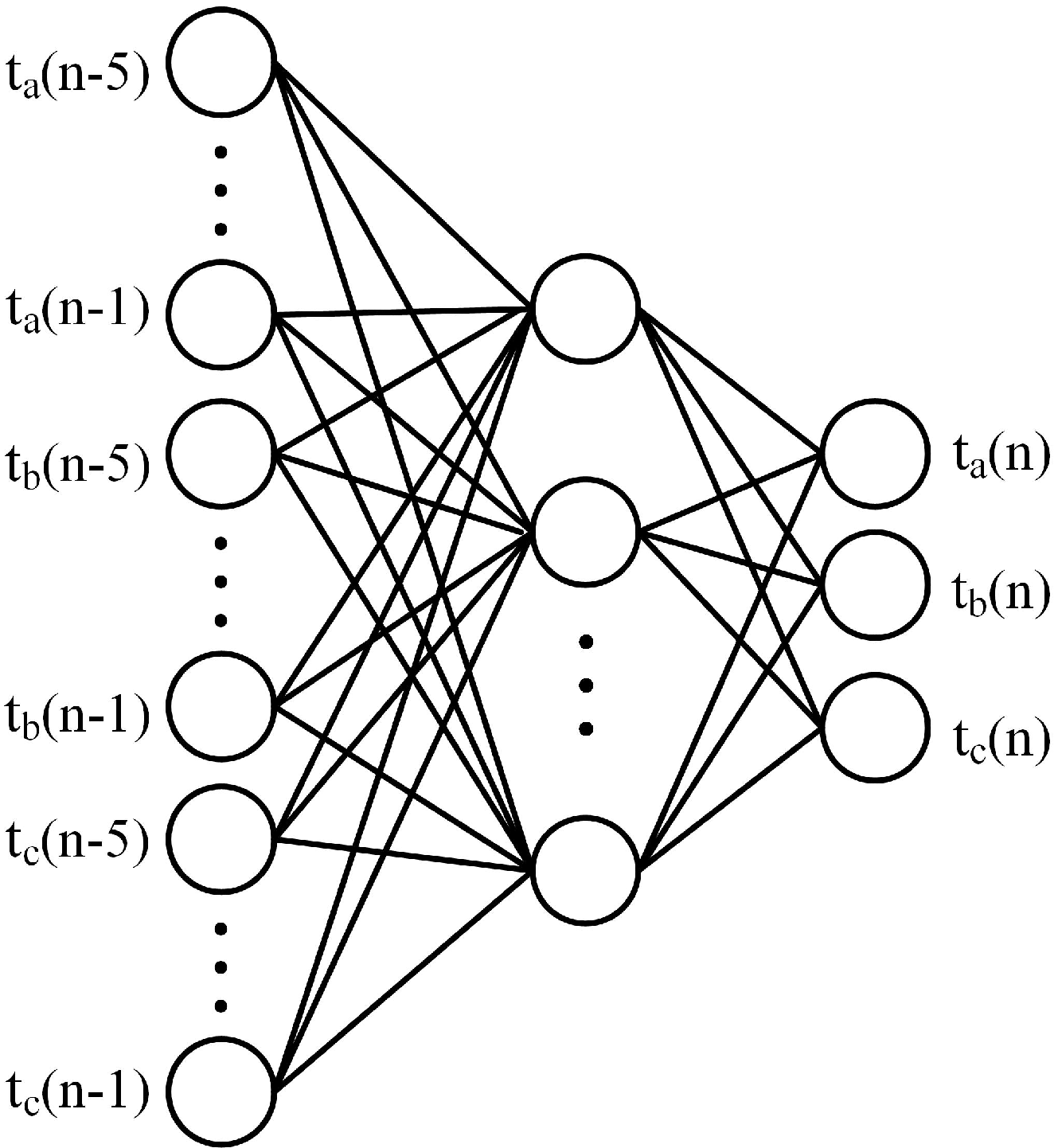}}
 \subfigure[Multi-link multi-task learning (MMTL) model]{
\includegraphics[scale=0.5]{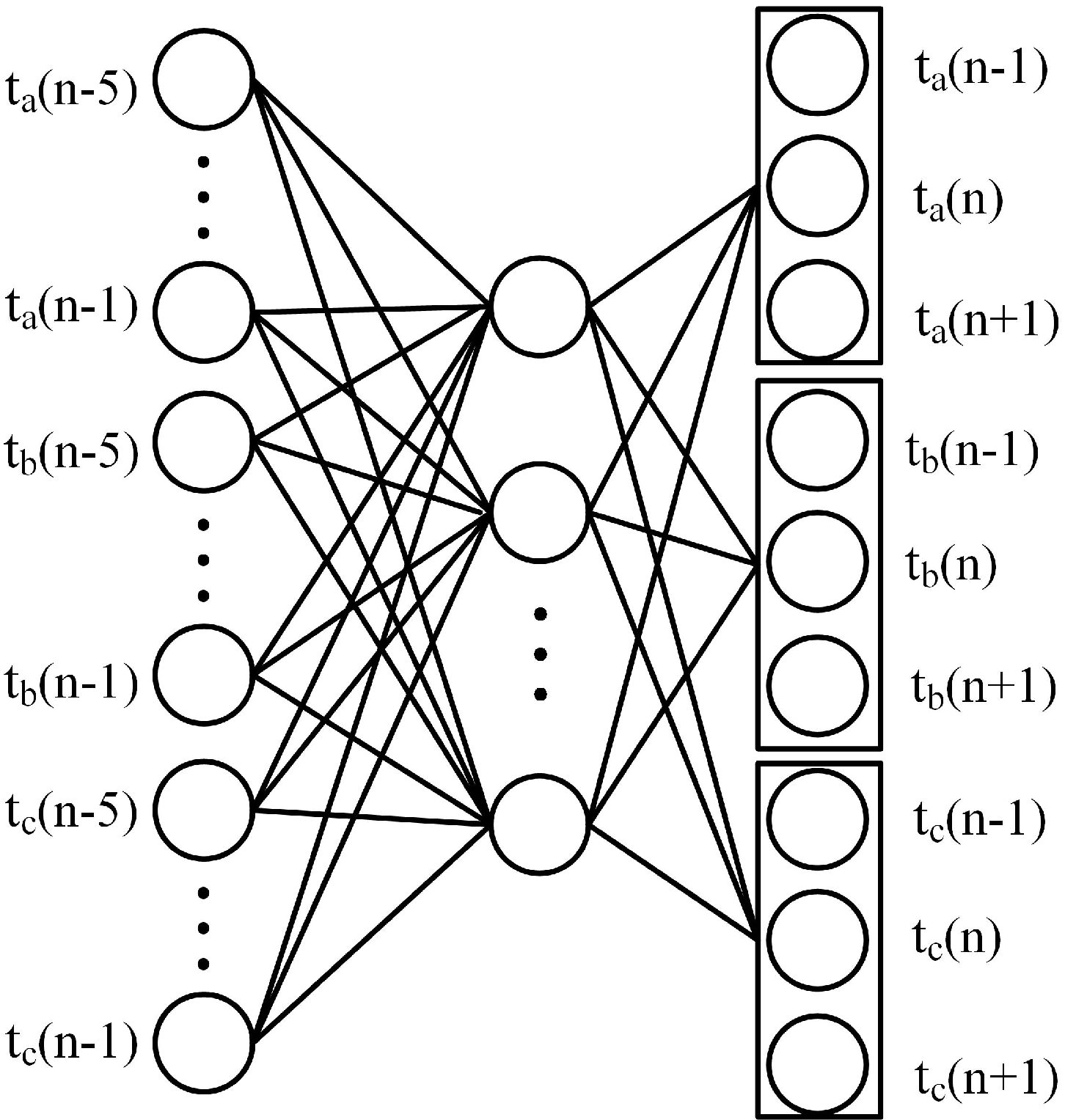}}
\caption{Diagrams of the two multi-link models based on NNs.}
\label{Figure 3}
\end{figure}

\section*{Gaussian Process Regression}
Gaussian process regression (GPR) is an important Bayesian machine
learning approach which places a prior distribution over the function space. All inferences of
GPR are taken place in the function space. In supervised learning
applications, it aims at the conditional distribution of the targets
given the inputs but not the distribution of the inputs. Moreover,
Gaussian process is a generalized Gaussian probability distribution.
The random variables in Gaussian probability distribution are
scalars or vectors (multivariate case). In Gaussian process
case, the random processes are represented as functions. There are several ways to
interpret GPR. Here we give a brief inference on the theoretical basis of GPR from the function-space
view. More details can be seen in~\citep{RasmussenGaussian06}.

Suppose that we have a training set $W=\{({\mathbf{x}}_i, y_i )| i= 1,
\dots, n\}$, where $n$ is the number of observations,
${\mathbf{x}}_i$ denotes the $i$-th D-dimensional input variable and
$y_i$ is the corresponding target which is always a real value in the
regression case. Aggregating the inputs as column vectors for all
$n$ case, we can get a $D\times n$ design matrix $X$.
Similarly, collect the targets in a vector as $\textbf{y}$. Then the
training set $W$ can be written as $W= (X, \textbf{y})$. In the
same way, we represent a test set $M=\{({\mathbf{x}}_i^*, y_i^* )| i=
1, \dots, n^* \}$ as $M= (X^*, {\textbf{y}}^*)$, where $X^*$ is a
$D\times n^*$ matrix.

A Gaussian process is specified by its mean function and
covariance function~\citep{RasmussenGaussian06}. Define the mean function $m({\mathbf{x}})$ and
covariance function $k({\mathbf{x}},{\mathbf{x'}})$ of a Gaussian
process $f({\mathbf{x}})$ as

\begin{equation}
\label{binFun}
\begin{array}{ll}
m(\textbf{x}) = {\textbf{E}}[f(\textbf{x})], \\
k(\textbf{x},\textbf{x}') = {\textbf{E}}[(f(\textbf{x}) - m(\textbf{x}))(f(\textbf{\textbf{x}}') - m(\textbf{x}'))] \; .
\end{array}
\end{equation}
Then, the Gaussian process can be written as
\begin{equation}
f(\textbf{x}) \sim \textbf{GP}(m(\textbf{x}),k(\textbf{x},\textbf{x}')).
\end{equation}

Generally, for the simplicity of notation and computation, the mean
function is set to be zero. In addition, Gaussian process has a
definition that a Gaussian process is a collection of random
variables and any finite number of them have a joint Gaussian
distribution. According to this definition, we can get a consistency
principle. That is, if there is a distribution $(y_1, y_2 ) \sim
N({\mathbf{\mu }}, \Sigma )$, then there exists $y_1\sim N(\mu _1 ,
\Sigma _{11} )$, where $\Sigma _{11}$ is the relevant submatrix of
$\Sigma$. This consistency principle plays an important role in the
inference of the GPR algorithm, which is also known as the marginalization
property.

Since prediction with noise-free observations is the special case
of prediction with noise observations, we take the noise case to do the inference of GPR. In the noise case, the relation
between the observed target value $y$ and the function value
$f({\mathbf{x}})$ is
\begin{equation}
y = \;f({\mathbf{x}}) + \varepsilon ,
\end{equation}
where $\varepsilon$ is the additive independent identically
distributed Gaussian noise with variance $\sigma _n^2$. In the GPR
inference, the covariance function should be predefined. We take the
squared exponential (SE) covariance function as an example.

\begin{equation}
\label{COV}
\begin{array}{ll}
  \operatorname{cov} (f(x_p ),f(x_q )) = k(x_p ,x_q )\\
   = \sigma _f \exp \left[ { - (x_p  - x_q )^T P^{ - 1} (x_p  - x_q )/2} \right],
\end{array}
\end{equation}
where $P$ is a diagonal matrix with $l_1, l_2, \ldots, l_D$ being the diagonal elements, $D$ is the dimension of the input
space, and $\sigma _f^2$ is the signal variance which controls the
global variation. From formula (\ref{COV}), we note that the
covariance between the outputs can be written as a function of the
inputs. Following the independence assumption and with the specified
covariance function, we can easily get the prior of the training set $X$. That is
\begin{equation}
 \label{COVY}
 \operatorname{cov} ({\mathbf{y}}) = K(X,X) + \sigma
_n^2 I.
\end{equation}
It is easy to find out that formula (\ref{COVY}) can also be seen as the
prior in noise-free case by removing the noise term. According to the
independence assumption, the noise term is a diagonal matrix.

With the prior, we can further get the prior joint distribution of
the observed target values $\mathbf{y}$ and the function values
${\mathbf{f}}^{\mathbf{*}} $ of the test samples:
\begin{equation}
\left[ {\begin{array}{*{20}c}
   {\mathbf{y}}  \\
   {{\mathbf{f}}^{\mathbf{*}} }  \\

 \end{array} } \right] \sim \,N\left( {0,\;\,\left[ {\begin{array}{*{20}c}
   {K(X,X) + \sigma _n^2 I} & {K(X,X^{\mathbf{*}} )}  \\
   {K(X^{\mathbf{*}} ,X)} & {K(X^{\mathbf{*}} ,X^{\mathbf{*}} )}  \\

 \end{array} } \right]} \right).
\end{equation}
To get the posterior distribution over functions, we need to reject
those functions that disagree with observations of the prior. In
probabilistic terms, this operation can be done easily. According
to properties of marginal distribution and conditional distribution~\citep{RasmussenGaussian06}, we can get
\begin{equation}
    \label{FD}
   \begin{array}{ll}
  f^* \left| {X,y,X} \right.^*  \sim \,N(\bar f^* ,\operatorname{cov} (f^* )),\mbox{where} \\
  \bar f^*  \triangleq E[f^* \left| {X,y,X} \right.^* ] = K(X^* ,X)[K(X,X) + \sigma _n^2 I]^{ - 1} y, \\
  \operatorname{cov} (f^* ) = K(X^* ,X^* )- K(X^* ,X)[K(X,X) + \sigma _n^2 I]^{ - 1} K(X,X^* ).  \\
\end{array}
\end{equation}
Formula (\ref{FD}) gives the distribution of the function values
${\mathbf{f}}^{\mathbf{*}}$. In practical applications, the mean
function value ${\mathbf{\bar f}}^*$ is evaluated as the output of
GPR. In fact, the GPR algorithm simultaneously outputs the variance of
prediction values, which is also considered as the potential
capability of GPR compared to other regression algorithms. A faster and more stable algorithm using Cholesky decomposition to compute
the inverse covariance matrix in formula (\ref{FD}) also can be found in~\citep{RasmussenGaussian06}.

\section*{Graphical Lasso}
Graphical lasso (GL) is an algorithm to construct a sparse graphical
model by applying the lasso penalty to the inverse covariance matrix. There is
a basic model assumption that the observations have a multivariate
Gaussian distribution with mean $\mu$ and covariance matrix $\Sigma$~\citep{Friedman08}. The key to build a sparse graphical model is to make
the inverse covariance matrix as sparse as possible. If the $ij$-th
component of $\Sigma ^{ - 1}$ is zero, then there is no link between
the two variables $i$ and $j$ in the sparse graphical model.
Otherwise, there exists a link between the two variables. In recent years, series of
approaches have been proposed to solve this problem. All the
approaches can be classified into two types: the approximate approaches and the exact
approaches. The approximate approaches estimate the sparse graphical
model by fitting a lasso model to each variable and using the others
as predictors~\citep{MeinshausenHighDimensional06}. The exact approaches solve the maximization
of the L1-penalized log-likelihood problem. There are also several ways
to solve the exact problem. For example, interior point
optimization methods~\citep{Dahl08} and blockwise coordinate descent (BCD)
algorithm~\citep{Friedman08}, etc. Thereinto, the BCD algorithm based on GL is appreciated
as a relatively efficient method~\citep{MeinshausenHighDimensional06}. Below, for completeness, we introduce the GL algorithm. More detailed information
can be found in related references.

\subsection*{Problem Setup}
Assume that we are given $N$ observations independently drawn from a
$p$-variate normal Gaussian distribution, with mean $\mu$ and
covariance $\Sigma$. Let $S$ denote the empirical covariance
matrix. Thus we have
\begin{equation}
\label{S}
 S = \frac{1} {N}\sum\limits_{k = 1}^N {(x_k  - \mu )}
(x_k  - \mu )^T ,
\end{equation}
where $x_k$ denotes the $k$-th observation. According to the
independence assumption, we can easily get the likelihood $L$ on the
given data set.
\begin{equation}
\begin{array}{ll}
  L = \mathop \prod \limits_{k = 1}^N p(x_k ;\Sigma )  \\
  \,\,\,{\kern 1pt} \, = \mathop \prod \limits_{k = 1}^N \{ \frac{1}
{{(2\pi )^{p/2} \left| \Sigma  \right|^{1/2} }}e^{ - (x_k  - \mu )^T \Sigma ^{ - 1} (x_k  - \mu )/2} \} \\
  \;\;\, = \frac{1}
{{(2\pi )^{Np/2} \left| \Sigma  \right|^{N/2} }}e^{ - \sum\limits_{k = 1}^N {(x_k  - \mu )^T \Sigma ^{ - 1} (x_k  - \mu )/2} }. \\
\end{array}
\end{equation}
Then, the log-likelihood can be written as
\begin{equation}
\label{LOGL}
\begin{array}{ll}
  \log L =  - \log [(2\pi )^{\frac{{Np}}
{2}} \left| \Sigma  \right|^{\frac{N}
{2}} ] - \sum\limits_{k = 1}^N {(x_k  - \mu )^T \Sigma ^{ - 1} (x_k  - \mu )/2}   \\
   =  - \frac{{Np}}
{2}\log 2\pi  - \frac{N} {2}\log \left| \Sigma  \right| - \frac{1}
{2}\sum\limits_{k = 1}^N {(x_k  - \mu )^T \Sigma ^{ - 1} (x_k  - \mu )} .  \\
\end{array}
\end{equation}
Because the GL algorithm has to solve the problem of maximizing the L1-penalized
log-likelihood, we make a few transformations on formula
(\ref{LOGL}). Removing the constant term in formula (\ref{LOGL}) and
combining it with formula (\ref{S}), we get
\begin{equation}
\begin{array}{ll}
  \log L \propto  - \log \left| \Sigma  \right| - \frac{1}
{N}\sum\limits_{k = 1}^N {(x_k  - \mu )^T \Sigma ^{ - 1} (x_k  - \mu )}  \hfill \\
   \propto \log \left| {\Sigma ^{ - 1} } \right| - trace(\frac{1}
{N}\sum\limits_{k = 1}^N {(x_k  - \mu )^T \Sigma ^{ - 1} (x_k  - \mu )} ) \hfill \\
   \propto \log \left| {\Sigma ^{ - 1} } \right| - trace(\Sigma ^{ - 1}  \cdot \frac{1}
{N}\sum\limits_{k = 1}^N {(x_k  - \mu )} (x_k  - \mu )^T ) \hfill \\
   \propto \log \left| {\Sigma ^{ - 1} } \right| - trace(\Sigma ^{ - 1}  \cdot S) \hfill \\
   \propto \log \left| {\Sigma ^{ - 1} } \right| - trace(S \cdot \Sigma ^{ - 1} ). \hfill \\
\end{array}
\end{equation}
Therefore, the exact problem that the GL algorithm solves can be written as
\begin{equation}
\label{GL}
 \Sigma ^{ - 1}  = \arg \mathop {\max }\limits_{X \succ
0} \log \left| X \right| - trace(S \cdot X) - \rho \left\| X
\right\|_1,
\end{equation}
where $\left\| X \right\|_1$ is the L1-norm of matrix $X$, which is
the sum of the absolute values of the elements of $X$, and $\rho$ is
the penalty parameter which controls the extent of
penalization~\citep{Banerjee08}.

\subsection*{Formula Transformations}
Focusing on formula (\ref{GL}), a series of
transformations are carried out to get an equivalent form which can be easily solved. Firstly,
we write the problem as
\begin{equation}
\label{GLA}
 \mathop {\max }\limits_{X \succ 0} \mathop {\min
}\limits_{\left\| U \right\|_\infty   \leqslant \rho } \log \left| X
\right| + trace(X,S + U),
\end{equation}
where $\left\| U \right\|_\infty$ denotes the maximum absolute value element
of the symmetric matrix $U$~\citep{Banerjee08}. Exchanging the max and min, formula~(\ref{GLA}) is transformed as follows.
\begin{equation}
\label{GLA13}
\mathop {\max }\log \left| X \right| - trace(X,S + U).
\end{equation}
Computing the derivative of formula~(\ref{GLA13}) over $X$, we
obtain $X={(S+U)}^{-1}$. Replace $X$ in formula~(\ref{GLA13}) with $(S+U)^{-1}$ and
it follows that the dual problem of formula~(\ref{GLA}) becomes
\begin{equation}
\mathop {\min }\limits_{\left\| U \right\|_\infty   \leqslant \rho }
- \log \left| (S + U) \right| - p,
\end{equation}
where $p$ is the dimension of matrix $X$, and the relation between the primal and the dual variables
is $X={(S+U)}^{-1}$. To write neatly, set $M=S+U$. Then, the dual
problem of the primal maximum L1-penalized log-likelihood problem is
\begin{equation}
\label{GLS}
 \Sigma  = \arg \max \{ \log \left| M \right|:\left\| {M - S}
\right\|_\infty \leqslant \rho \}.
\end{equation}
According to the series of transformations above, we find that we finally
estimate $\Sigma$ in the dual problem~(\ref{GLS}) while the inverse
covariance matrix $\Sigma ^{ - 1}$ in the primal problem~(\ref{GL}).
Moreover, we also observe that the diagonal elements of $\Sigma$ and
$S$ have a relation as
\begin{equation}
\Sigma _{ii}  = S_{ii}  + \rho,
\end{equation}
which holds for all $i$.

\subsection*{Block Coordinate Descent (BCD) Algorithm}
Let $W$ be the estimation of $\Sigma$. The block coordinate descent
(BCD) algorithm solves problem~(\ref{GLS}) by optimizing
cyclically over each row and column of $W$, until achieving the given
convergence condition. More details about BCD algorithm can be seen
in~\citep{Banerjee08}. As to the GL approach we discussed, BCD algorithm plays as a
launching point.

Divide $W$ and $S$ into blocks as
\begin{equation}
W = \left( {\begin{array}{*{20}c}
   {W_{11} } & {w_{12} }  \\
   {w_{12}^T } & {w_{22} }  \\

 \end{array} } \right),S = \left( {\begin{array}{*{20}c}
   {S_{11} } & {s_{12} }  \\
   {s_{12}^T } & {s_{22} }  \\

 \end{array} } \right).
\end{equation}
The BCD algorithm updates $w_{12}$ through solving the quadratic  programming

\begin{equation}
\label{W12}
 w_{12}  = \arg \min _y \{ y^T W_{11}^{ - 1} y:\left\| {y
- s_{12} } \right\|_\infty   \leqslant \rho \},
\end{equation}
which is solved by an interior point procedure. Permuting the rows
and columns to make the target column always be the last one, BCD
solves the problem like formula~(\ref{W12}) for each column
and updates its estimation for $W$ when all columns have been processed. This process is repeated until convergence.

The dual problem of formula~(\ref{W12}) shown as the
following is also deduced in~\citep{Banerjee08}.
\begin{equation}
\label{Beta}
 \min _\beta  \{ \frac{1} {2}\left\| {W_{11}^{1/2} \beta
- b} \right\|^2  + \rho \left\| \beta  \right\|_1 \},
\end{equation}
where $b = W_{11}^{ - 1/2} s_{12}$. If $\beta$ solves formula~(\ref{Beta}), then the solution of formula~(\ref{W12}) is
$w_{12}=W_{11} \beta.$ It can be easily found that formula~(\ref{Beta})
resembles a lasso regression problem, which is the launching
point of GL approach. There is a verification on the
equivalence between the solutions of formula~(\ref{GL}) and (\ref{Beta}) given in~\citep{Friedman08}.

\subsection*{Algorithm Description and Realization}
According to the lasso problem~(\ref{Beta}) achieved by the BCD algorithm,
the GL approach solves and updates this problem recursively. Details of
the GL algorithm can be described as

1. Set $W= S+ \rho I$, where $I$ is the identity matrix. Then the
diagonal of $W$ remains unchanged in all the following steps.

2. For each row and column of $W$, solve the lasso problem~(\ref{Beta}) and obtain the solution $\beta$.

3. Compute $w_{12}$ by $w_{12}  = W_{11} \beta$, and replace the corresponding row and column of $W$ with $w_{12}$.

4. Repeat the above steps 2 and 3 until convergence.

5. Compute the inverse matrix of $W$, which is also the required inverse covariance matrix $\Sigma ^{ - 1}$.

In~\citet{Friedman08}, there also gives a relatively cheap method to execute the
last step of the GL algorithm above. From the achieved
sparse matrix $\Sigma ^{ - 1}$, GL builds the desired sparse
undirected graphical model. Each row or column of matrix $\Sigma ^{
- 1}$ represents a node in the graphical model. The row or column is
corresponding to a variable of the multi-variable data. Therefore,
$p$-dimensional data has $p$ nodes in the graphical model. Whether
there is a link between two nodes is determined by the
corresponding component of matrix $\Sigma ^{ - 1}$ being zero or not.
If the component is zero, then there is no link in the graphical
model. That is, the two variables are conditionally independent given other
variables. In the next section, combing with the multi-link traffic flow
prediction model, we give an instance of building the sparse graphical
model by GL.

\section*{Experiments}
\subsection*{Data Description}
The data sets used in this paper are vehicle flow rates recorded
every 15 minutes, which were gathered along many road links by the
UTC/SCOOT system of Beijing Traffic Management~\citep{Sun07}. The unit
of the data is normalized as vehicles per hour (vehs/h). For the
short-term traffic flow forecasting, we carry a one-step prediction
and take 15 minutes as the prediction horizon. That is, we predict
the traffic flow rates of the next 15-minute interval every time.

From the urban traffic map, we select a portion including 31 road
links shown as Fig.~1. Each circle node in the figure represents a
road junction which combines several road links. The arrows show the
directions of traffic flows from the upstream junctions to the
corresponding downstream junctions. Paths without arrows denote no
traffic flow records. Raw data are taken from March 1 to March 31,
2002, totally 31 days. Considering malfunctions of traffic flow
detectors, we wiped away the days with empty data. Finally, the
remaining data we used are of 25 days and have totally 2400 sample points.
We divide the data into two parts, the first 2112 samples as
training data and the rest as test data.

\subsection*{Model Building with GL}
In multi-link traffic flow prediction case, we can use certain
historical traffic flows of all the links in the whole map. Through
building a sparse graphical model by GL, we extract the informative historical
traffic flows provided by all the links. Basing on the data set
described above, we take 6 continuous traffic flows of each link to
build the sparse graphical model. Because there are 31 links, we
will get a $186\times186$ inverse variance matrix $(6\times31=186)$.
For traffic flow forecasting, the first 5 historical traffic flows of 31 links are all used to
predict the 6-th traffic flow of one link. As to a predicted link, when building the graphical model,
we need the 6 traffic flows on the predicted link and the other 30 links' first 5 historical traffic flows.
Therefore, to a certain link, there are at most 186-30=156 nodes in the graphical model.

Still take link Ba as an example. In the single-link prediction model, we
predict the traffic flow Ba(n) using the continuous historical
traffic flows Ba(n-5), $\ldots$, Ba(n-1) on link Ba. While in the
multi-link prediction model, we consider the historical traffic
flows of all adjacent links or all the links in the whole traffic
map. The latter one seems to be more comprehensive, but it also brings
too much computation. Fortunately, this problem can be easily solved by
the GL approach. With the sparse graphical model built by GL,
we can extract the most relevant historical traffic flows to predict the predicted traffic flow.
In modeling of link Ba, we just consider the components of the corresponding column
or row in the inverse covariance matrix. If the component is zero, it means there
is no relevance or very little relevance between the two variables. Then there is no link
between the two corresponding variables in the graphical model. For example, let variable $i$
represent the predicted traffic flow and variable $j$ represent some historical traffic flow.
If there is no link between variables $i$ and $j$, it means historical traffic flow $j$ contributes nothing to the prediction of traffic flow $i$.
Fig.~4 gives the sparse graphical model of link Ba built by GL.

\begin{figure}[thb]
\centering
\includegraphics[scale=0.5]{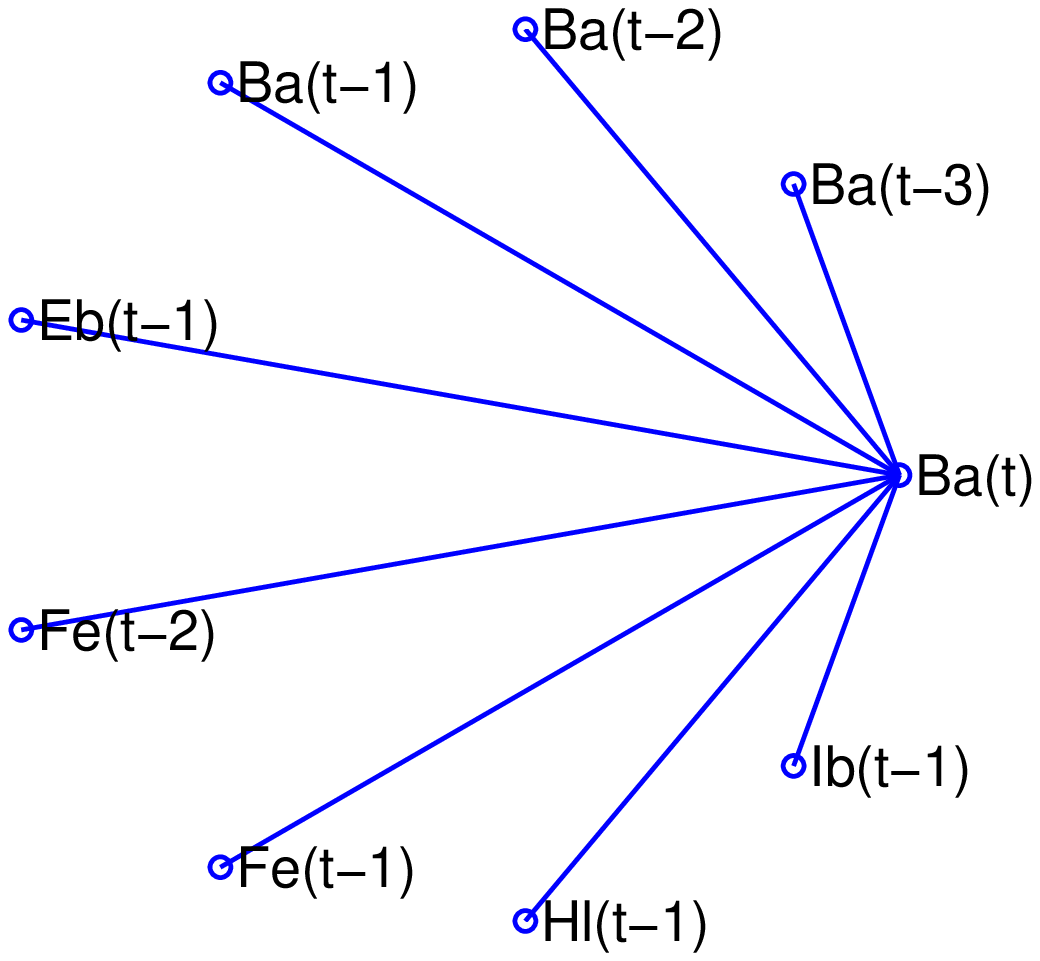}
 \caption{The sparse graphical model of link Ba built by GL.}
 \label{Figure 4}
\end{figure}

By comparing Fig.~1 and Fig.~4 with link Ba, we can see that the prediction of link Ba is not only relevant to the three traffic
flows Ba(t-3), Ba(t-2), Ba(t-1) on link Ba itself but also the other five traffic flows Eb(t-1), Fe(t-2), Fe(t-1) and Hl(t-1), Ib(t-1) of
link Eb, Fe, Hl and Ib respectively. There are only 8 variables considered relevant to the prediction of Ba(t), which is much fewer
than all the 155 variables considered in the general multi-link model. Therefore, GL further extracts the relevant information based on
our previous multi-link prediction model.

\subsection*{Experimental Settings}
In the design of NNs, a three-layer BP neural network is selected.
On the one hand, a three-layer NN can approximate arbitrary bounded
and continuous functions~\citep{Duda01}, and on the other hand, more
layers will make the network more complex.
Besides, BP NNs are well known for their good self-learning capability.
The number of input and output units is determined by the dimension of the experimental data.
For example, in single-link models based on NNs, as we use 5 historical traffic flows to predict the traffic flow of the next time interval,
the number of input units is 5, while the number of output units is 1 in the SSTL model and 3 in the SMTL model. The
case of the multi-link model also can be inferred from the representation of the multi-link model in the section of multi-link model. In GL with NN
case, the number of input units depends on the extracted dimension
of the GL algorithm and the number of output units is the same as the SSTL
model. Obviously, different links will have different numbers of input
units in GL with the NN case. For all approaches based on NNs, the number
of hidden units is computed by the empirical formula shown below.
\begin{equation}
n = \sqrt {n_i  + n_0 }  + a,
\end{equation}
where $n$, $n_i$ and $n_0$ respectively denote the number of hidden,
input and output-layer units, $a$ is a constant that can be chosen
between 1 and 10\citep{ZhangMultiple10}. To obtain a relatively optimal construction of NN,
we try different values of $a$ from 1 to 10 with an interval 1 and finally
choose the one with best performance. As to the transfer
functions of NN, \emph{sigmoid} function is selected between the input layer
and the hidden layer, \emph{purelin} function is selected between the hidden
layer and the output layer. The \emph{trainlm} function is selected as the
training function, because it is based on Levenberg-Marquardt algorithm
and can converge rapidly with a high prediction accuracy.

In the realization of GPR, we need to specify the covariance
function and find a method to optimize the parameters. In our
experiments, we choose the following squared exponential (SE)
covariance function.

\begin{equation}
k(\textbf{x}_p ,\textbf{x}_q ) = \sigma _f^2 \exp \left[ { -
(\mathbf{x}_p  - \mathbf{x}_q )^{T} {P}^{ - 1}
(\mathbf{x}_p  - \mathbf{x}_q )/2} \right],
\end{equation}
where $P$ is a diagonal matrix with diagonal elements being $l_1, l_2,
\dots, l_D$, $D$ is the dimension of the input space, $\sigma _f^2$ is
the signal variance which controls the global variation. Therefore,
together with the noise variance $\sigma _n^2$ involved in GPR
algorithm, there are totally $D+2$ parameters. Following the
suggestion in~\citep{RasmussenGaussian06}, we initially set $l_1, l_2, \dots, l_D$, $\sigma
_f^2$ all as 1 and $\sigma _n^2$ as 0.1. Then, we optimize these
parameters by maximizing the marginal likelihood
\begin{equation}
\log p(\mathbf{y}|X) =  - \frac{1} {2}{\mathbf{y}}^{\mathbf{T}}(K + \sigma _n^2 I)^{ - 1} {\mathbf{y}}- \frac{1} {2}\log \left| {K + \sigma _n^2 I}\right| - \frac{n} {2}\log 2\pi ,
\end{equation}
which can also be easily achieved by $y \sim N(0,K + \sigma _n^2
I)$. We use the gradient descent algorithm to minimize the
negative marginal likelihood to get the optimal parameters.

Similarly, there are two parameters to be specified in GL. One is the penalty parameter $\rho$, and the other is the
lower limit value which determines the relevance between two
variables. As to the selection of the penalty
parameter, we follow the suggestion in Section 2.3 of~\citep{Banerjee08}.
There is also a statement saying that, if we follow the suggestion
to choose the penalty parameter, the error rate of estimating the
graphical model can be controlled. Since the GL algorithm builds the
sparse graphical model according to the inverse covariance matrix,
we need a lower limit value to screen out the nonzero
components of the inverse covariance matrix that represent effective information. In
our experiments with GL, we set the component of the inverse
covariance matrix as zero if it is less than 5e-4. That is, we think that
there is little relevance between the two variables when the
corresponding component in the inverse covariance matrix is so
small.

\subsection*{Results}
To examine the rationality of the multi-link model, we first compute the correlation coefficients
of adjacent links. There are 10 junctions in the traffic map of Fig.~1, and there are
totally 36 correlation coefficients. We list the 36 correlation coefficients in Table~\ref{tab1}.
\begin{table}[thb]
 \caption{The correlation coefficients of adjacent links corresponding to 10 junctions.}
 \label{tab1}
\begin{center}
 \begin{tabular}{|c|c|c|c|c|c|c|}
  \hline Junction&    cor\_coef   &      cor\_coef  &      cor\_coef  &    cor\_coef    &   cor\_coef   &  cor\_coef  \\
  \hline B       &    0.9512      &      0.9370     &      0.9608     &                 &               &             \\
  \hline C       &0.9320&0.8731&0.8836&0.9446&0.9510&0.9365\\
  \hline D       &0.7998&0.9073&0.9218&0.7405&0.7493&0.9398\\
  \hline E       &0.7869&&&&&\\
  \hline F       &0.9420&0.8875&0.9637&0.9019&0.9359&0.9289\\
  \hline G       &0.9510&&&&&\\
  \hline H       &0.9423&0.9597&0.9473&&&\\
  \hline I       &0.8735&0.8987&0.9543&&&\\
  \hline J       &0.9238&&&&&\\
  \hline K       &0.8708&0.7280&0.7876&0.6662&0.7683&0.5988\\
  \hline
 \end{tabular}
 \end{center}
\end{table}
From Table~\ref{tab1}, we can see that the minimum, the maximum and the mean of the 36 correlation coefficients are
0.5988, 0.9637 and 0.8790, respectively. Thereinto, almost all of the 36 correlation coefficients are larger than 0.8,
which means high correlation between variables. Therefore, as to the real-world data sets we used in the experiments,
it is reasonable and meaningful to test the proposed multi-link approaches on them.

In this paper, the mentioned approaches for traffic flow forecasting include SSTL, SMTL, MSTL, MMTL, GPR and GL$\_$NN, respectively.
We test them on the 31 real-world traffic flow data sets described above, which are collected from the 31 road links of Fig.~1.
To get a complete evaluation of all the proposed approaches, the historical average, which we mark as Hist\_Avg, is adopted as a base line for comparison.
We adopt root mean square error ($RMSE$) and mean absolute relative error ($MARE$) to evaluate the
prediction performance of different approaches. $RMSE$ and $MARE$ are formulated as follows.
\begin{equation}
\label{RMSE}
RMSE{=}\sqrt {\frac{1} {N}\sum\limits_{i = 0}^N {(t(i) -t'(i))^2 } } ,
\end{equation}
and
\begin{equation}
\label{MARE}
MARE{=}\frac{1} {N}\sum\limits_{i = 0}^N {\frac{|t(i) -t'(i)|}{t(i)}}  ,
\end{equation}
where $t'(i)$ is the prediction of $t(i)$, and $N$ is the number of test samples.

In Table~\ref{tab2} and Table~\ref{tab3}, we present the experimental results of
$MARE$ and $RMSE$ on the 31 road links corresponding to all the compared approaches.
\begin{table}[thb]
 \caption{$MARE$s(\%) of the 31 links corresponding to all the compared approaches.}
 \label{tab2}
 \begin{center}
 \begin{tabular}{|c|c|c|c|c|c|c|c|}
  \hline MARE &GPR& SSTL & SMTL & MSTL  & MMTL  &GL\_NN & Hist\_Avg\\
  \hline Ba &11.55 &12.83  &11.14 &10.61  & 11.43 &11.28  &12.94\\
  \hline Bb &7.72  &7.98   &8.02  & 7.57  & 7.78  &7.68   &9.01 \\
  \hline Bc &9.34  &9.95   &9.40  & 8.38  & 8.23  &8.26   &9.90 \\
  \hline Ce &9.82  &10.30  &10.23 &9.34   & 9.53  &9.52   &9.04 \\
  \hline Cf &8.92  &9.03   &8.95  &9.08   & 8.32  &7.58   &8.76 \\
  \hline Cg &11.86 &12.53  &12.00 &11.45  & 11.67 &13.37  &14.70\\
  \hline Ch &10.56 &10.18  &10.61 &10.19  & 10.12 &9.47   &9.74\\
  \hline Da &20.85 &20.51  &18.89 &18.44  &	17.16 &20.02  &23.79\\
  \hline Db &24.61 &25.02  &24.73 &25.06  &	24.68 &26.26  &21.11\\
  \hline Dc &13.31 &13.95  &13.84 &16.79  &	14.59 &12.20  &12.97\\
  \hline Dd &12.94 &13.17  &13.50 &12.52  &11.80  &9.94   &12.39\\
  \hline Eb &10.83 &12.74  &11.13 &12.92  &12.74  &10.24  &14.34\\
  \hline Ed &15.51 &16.35  &15.36 &16.66  &16.16  &14.21  &30.04\\
  \hline Fe &7.74  &8.38   &7.50  &7.94   &10.07  &7.54   &9.81\\
  \hline Ff &11.29 &11.68  &11.81 &10.65  &11.32  &12.50  &13.97\\
  \hline Fg &10.08 &11.20  &9.88  &9.69   &11.10  &9.34   &11.14\\
  \hline Fh &8.52  &10.02  &9.31  &9.11   &9.39   &8.69   &9.81 \\
  \hline Gb &13.37 &14.10  &15.30 &13.52  &12.80  &12.80  &14.52\\
  \hline Gd &10.36 &11.11  &10.63 &10.22  &9.96   &8.84   &11.83\\
  \hline Hi &11.64 &11.79  &11.58 &12.34  &11.27  &12.12  &15.48\\
  \hline Hk &13.44 &14.03  &14.28 &12.78  &12.96  &13.33  &16.60\\
  \hline Hl &9.29  &9.79   &10.07 &9.15   &9.53   &8.00   &10.71\\
  \hline Ia &16.00 &16.85  &16.66 &15.88  &15.90  &18.20  &20.86\\
  \hline Ib &9.77  &9.51   &9.39  &8.71   &8.60   &8.04   &8.53\\
  \hline Id &8.07  &8.46   &8.37  &8.37   &8.61   &6.92   &8.14\\
  \hline Jh &7.95  &8.02   &7.89  &8.29   &8.69   &8.26   &8.47\\
  \hline Jf &9.23  &9.43   &9.66  &7.88   &7.61   &6.92   &10.66\\
  \hline Ka &9.24  &9.50   &9.23  &10.00  &10.24  &8.51   &10.86\\
  \hline Kb &10.31 &10.48  &10.73 &10.30  &10.08  &10.69  &11.38\\
  \hline Kc &26.01 &28.51  &31.39 &27.63  &31.14  &25.00  &27.54\\
  \hline Kd &11.80 &11.81  &11.54 &11.06  &11.26  &10.31  &14.52\\
  \hline
 \end{tabular}
 \end{center}
\end{table}

\begin{table}[thb]
 \caption{$RMSE$s of the 31 links corresponding to all the compared approaches.}
 \label{tab3}
\begin{center}
 \begin{tabular}{|c|c|c|c|c|c|c|c|}
  \hline RMSE &      GPR  &       SSTL  &      SMTL  &      MSTL  &      MMTL   &       GL\_NN  &Hist\_Avg\\
  \hline Ba &     \textbf{142.76} &     148.99 &     \textbf{147.16} &     150.81 &     \textbf{147.79} &       \textbf{139.71}&174.76\\
  \hline Bb &     \textbf{67.80}  &     \textbf{72.15} &     \textbf{71.86} &      73.60  &     72.59 &      \textbf{70.46}  & 85.79\\
  \hline Bc &     \textbf{96.83}  &     104.11 &     103.56 &     \textbf{98.80} &     \textbf{97.65} &      \textbf{91.89}&123.11\\
  \hline Ce &     \textbf{51.95}  &     55.65  &     55.31 &      \textbf{54.73}  &     \textbf{53.57}  &        \textbf{52.70}&52.53\\
  \hline Cf  &     91.34 &      89.31 &      88.87 &      \textbf{86.79} &      \textbf{84.58} &        \textbf{81.62}&105.06\\
  \hline Cg &     \emph{50.87} &      50.32 &     50.56 &      \textbf{49.51} &      \textbf{49.19 } &       53.27 &69.12\\
  \hline Ch &     67.35  &     65.95 &      66.01 &      \textbf{63.48} &     \textbf{63.13}  &       64.02 &67.20\\
  \hline Da &     112.40 &     \textbf{77.44} &      79.05 &      82.28 &      \textbf{77.15} &       95.47&132.70\\
  \hline Db &     \textbf{50.81} &      \textbf{53.29} &      \textbf{53.24} &      54.60 &      53.49 &      63.75&60.73\\
  \hline Dc &     \textbf{78.49} &      \textbf{85.93} &      \textbf{85.88} &      88.32 &      87.69  &     \textbf{ 73.39}&81.15\\
  \hline Dd &     \emph{62.93} &      \textbf{62.08 } &      \textbf{61.60} &      68.61  &     65.07 &     \textbf{ 55.99}&80.07\\
  \hline Eb &     \textbf{154.78} &     166.89 &     \textbf{162.26} &     168.14 &     \textbf{165.58} &      \textbf{150.17}&212.08\\
  \hline Ed &     195.81 &     \textbf{191.85} &     \textbf{196.43} &     208.95 &     199.36 &      \textbf{179.67}&340.79\\
  \hline Fe &     \textbf{116.60} &     116.80 &      115.69 &     \textbf{122.73} &     \textbf{119.94} &      \textbf{112.40}&160.03\\
  \hline Ff &     87.47 &      84.62 &      84.74 &      \textbf{83.88} &     \textbf{83.23}  &      103.15&106.47\\
  \hline Fg &     \textbf{85.10} &      95.16 &      \textbf{92.85} &      93.12 &      \textbf{92.40}  &      \textbf{87.67}&108.79\\
  \hline Fh &     \emph{151.53} &     151.51 &     149.71 &     \textbf{141.46} &     \textbf{136.23} &      144.00&171.69\\
  \hline Gb &     \emph{85.59} &      85.25 &      84.77 &      \textbf{83.64} &      \textbf{83.34} &      103.03&102.68\\
  \hline Gd &     157.37 &     \textbf{152.95} &     \textbf{151.42} &     153.39 &     155.08 &      \textbf{144.28}&191.14\\
  \hline Hi &     \emph{90.29} &      89.54 &      88.50  &     \textbf{87.23} &      \textbf{87.10} &      95.11&128.12\\
  \hline Hk  &     149.57 &     137.16 &     140.78 &     \textbf{131.72} &     \textbf{131.61} &      158.27&175.22\\
  \hline Hl &     \textbf{130.24} &     132.59 &      \textbf{129.23} &     \textbf{130.04} &     129.67 &      \textbf{108.92}&144.20\\
  \hline Ia &     \textbf{83.22} &      \textbf{86.54}  &     \textbf{86.10} &      88.60 &     88.13  &      100.65&118.84\\
  \hline Ib &     140.21 &     136.05 &     135.40 &     \textbf{132.83} &     \textbf{129.45} &      \textbf{124.16}&136.44\\
  \hline Id  &     \textbf{122.52} &     \textbf{134.42} &     134.45 &     135.06 &     \textbf{133.13} &      \textbf{113.36}&125.31\\
  \hline Jh &     119.73 &     \textbf{118.34} &      \textbf{116.65} &     148.23 &     148.88 &      \textbf{130.23}&136.04\\
  \hline Jf &     \textbf{137.57} &     159.15 &     160.30 &     120.33 &     \textbf{ 119.46} &      \textbf{108.42}&171.17\\
  \hline Ka &     81.38 &      77.31  &     77.07  &     75.72 &      \textbf{76.45}  &      \textbf{75.60}&96.70\\
  \hline Kb  &     146.35 &     142.16 &     141.15 &     134.27 &     \textbf{130.85} &      159.13&160.19\\
  \hline Kc &     \textbf{371.17} &     \textbf{384.20} &     \textbf{382.86} &     385.35 &     378.47 &      365.17&410.77\\
  \hline Kd &     172.14 &     168.21 &     167.22 &     163.50  &     \textbf{161.21} &      \textbf{159.61}&218.94\\
  \hline
 \end{tabular}
 \end{center}
\end{table}

In order to get a comprehensive comparison, we compare the experimental results from two views: the global and the local.
Table~\ref{tab2} and Table~\ref{tab3} locally give the $MARE$s and the $RMSE$s of the 31 links corresponding to all
the compared approaches. Fig.~5 globally shows the sum of the $RMSE$s of the 31 links corresponding to all the compared approaches.
\begin{figure}[thb]
\centering
\includegraphics[scale=0.8]{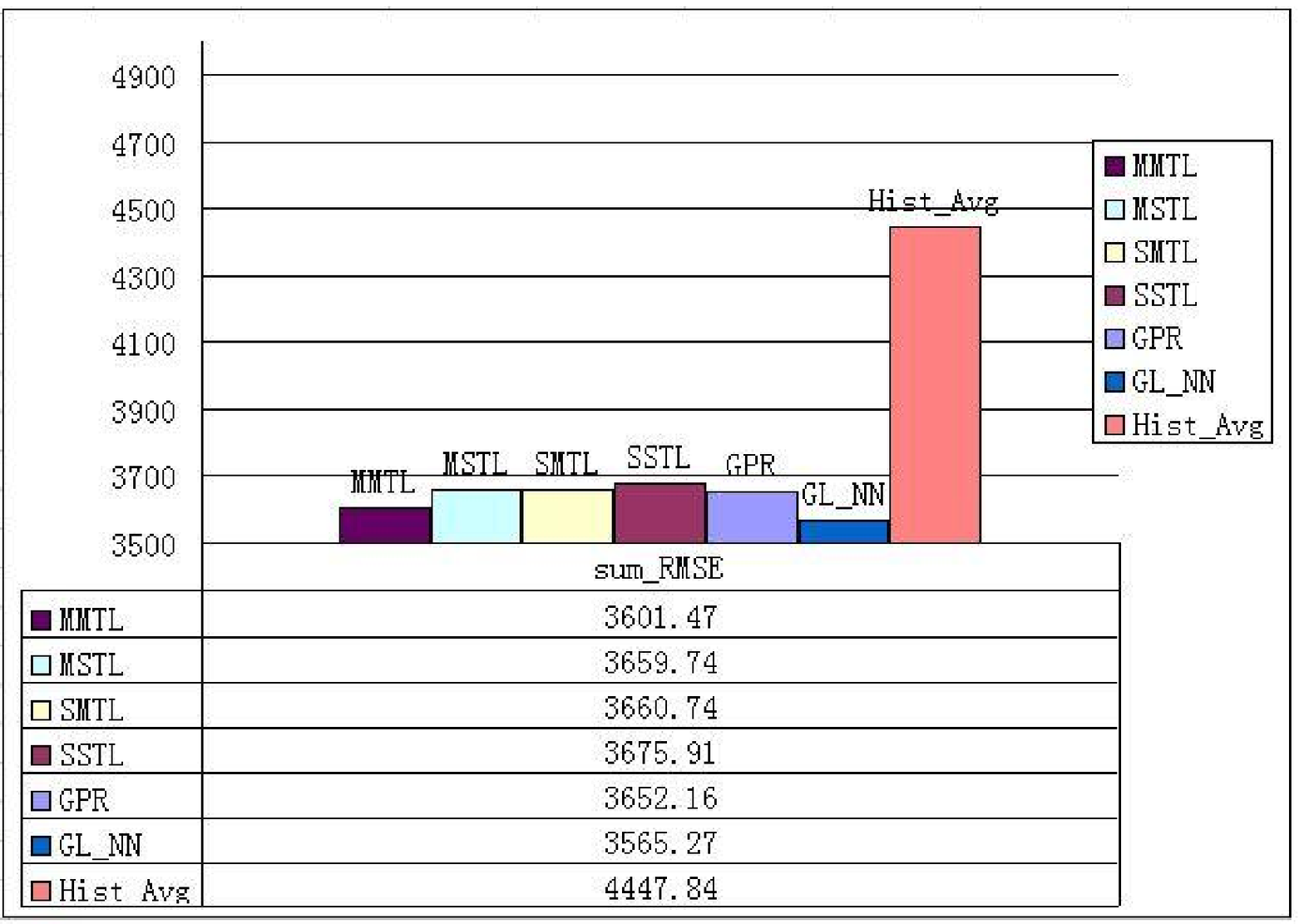}
 \caption{Sum of the $RMSE$s of the 31 links corresponding to all the compared approaches.}
 \label{Figure 5}
\end{figure}

Besides these, we further compare all the proposed methods using t-test. The t-test results of all the compared approaches are listed on Table~\ref{tab4}.
\begin{table}[thb]
 \caption{T-test(p value) results of all the compared approaches.}
 \label{tab4}
\begin{center}
 \begin{tabular}{|c|c|c|c|c|c|c|c|}
  \hline approaches& SSTL   &  SMTL    &  MSTL   &    MMTL    &   GPR   &  GL\_NN  &Hist\_Avg  \\
  \hline SSTL      &        &0.2074    &  0.0036 &    0.0226  &   0     &   0      &0.0394      \\
  \hline SMTL      &        &          &  0.1231 &    0.1792  &   0.0838&   0.0052 &0.0302      \\
  \hline MSTL      &        &          &         &    0.6961  &   0.9214&   0.0615 &0.0047      \\
  \hline MMTL      &        &          &         &            &   0.7299&   0.0612 &0.0095      \\
  \hline GPR       &        &          &         &            &         &   0.0183 &0.0035      \\
  \hline GL\_NN    &        &          &         &            &         &          &0.0005      \\
  \hline Hist\_Avg &        &          &         &            &         &          &            \\
  \hline
 \end{tabular}
 \end{center}
\end{table}

Firstly, we compare all the proposed approaches with the base line Hist\_Avg. According to the experimental results shown
in Table~\ref{tab2}, Table~\ref{tab3} and Fig.~5, of the 31 data sets, the number of data sets on which SSTL, SMTL, MSTL, MMTL, GPR and GL\_NN performs better than Hist\_Avg is
19, 21, 22, 22, 24 and 29, respectively, which means that all the proposed traffic flow forecasting approaches are superior to Hist\_Avg.
Secondly, we compare the performance of all the proposed approaches with each other such as the comparison of single-link approaches with multi-link approaches and
single-task approaches with multi-task approaches. GPR is a single-link single-task prediction approach in nature, when evaluating its performance, we compare it
with SSTL. As to the comparison criterion $MARE$, GPR outperforms SSTL on 28 data sets according to Table~\ref{tab2}. And according to t-test results shown in Table~\ref{tab4},
we can find that GPR is significantly better than SSTL. In the GPR column of Table~\ref{tab3}, we marked the components in bold which show that GPR is better than SSTL. We can see that, in the total
31 links, there are 14 links showing that GPR outperforms SSTL. However, there are another 5 links (italics in the GPR column of Table~\ref{tab3}) with error rate difference of GPR and SSTL less than 1. In Fig.~5, GPR is globally better than SSTL. Therefore, we still can conclude that GPR outperforms SSTL in traffic flow forecasting. According to the experimental results based on $RMSE$ in Table~\ref{tab3}, in the columns corresponding to SSTL, SMTL, MSTL, MMTL, we marked the two best components of the four approaches in bold. The numbers of boldfaces
are respectively 11, 13, 17 and 21. Therefore, we can get the conclusion that, multi-link approaches perform better than single-link approaches and multi-task learning approaches are better than single-link learning approaches. This is also why MMTL performs best in the four approaches. GL$\_$NN constitutionally belongs to the multi-link single-task prediction approaches.
The difference between GL$\_$NN and MMTL is that GL$\_$NN extracts the relevant information. From Table~\ref{tab2}, GL\_NN performs better than MMTL on 22 data sets. In the GL$\_$NN column of
Table~\ref{tab3}, we marked the components in bold which show that GL$\_$NN is better than MMTL. It is easy to find out that there are 21 links of all 31
links showing that GL$\_$NN is better than MMTL. According to the t-test result of GL\_NN and MMTL in Table~\ref{tab4}, it shows that GL\_NN outperforms MMTL to some extent.
The results fully verify the superiority of GL used in extracting information through building the sparse graphical model. GL\_NN performs best on traffic flow forecasting compared to all
the other proposed approaches.

\subsection*{Discussions on GPR}
As a single-link single-task approach, GPR performs better than SSTL on traffic flow forecasting.
In this section, we would give special illustrations of GPR on the potential capability.
Actually, the GPR algorithm outputs two terms that are the mean and the variance to be predicted.
Exactly speaking, GPR gives the distribution of the targets rather than the exact values. When computing the prediction errors, we use the mean as the prediction value.
Take link Kd as an example, Fig.~6 shows the practical prediction results. The star curve represents the actual values while the dot curve represents
the prediction values. The shaded area is the fluctuating range of targets predicted by GPR. As we can see, with the shaded part,
more actual targets can be contained in the prediction scope.

\begin{figure}[thb]
\centering
\includegraphics[scale=0.35]{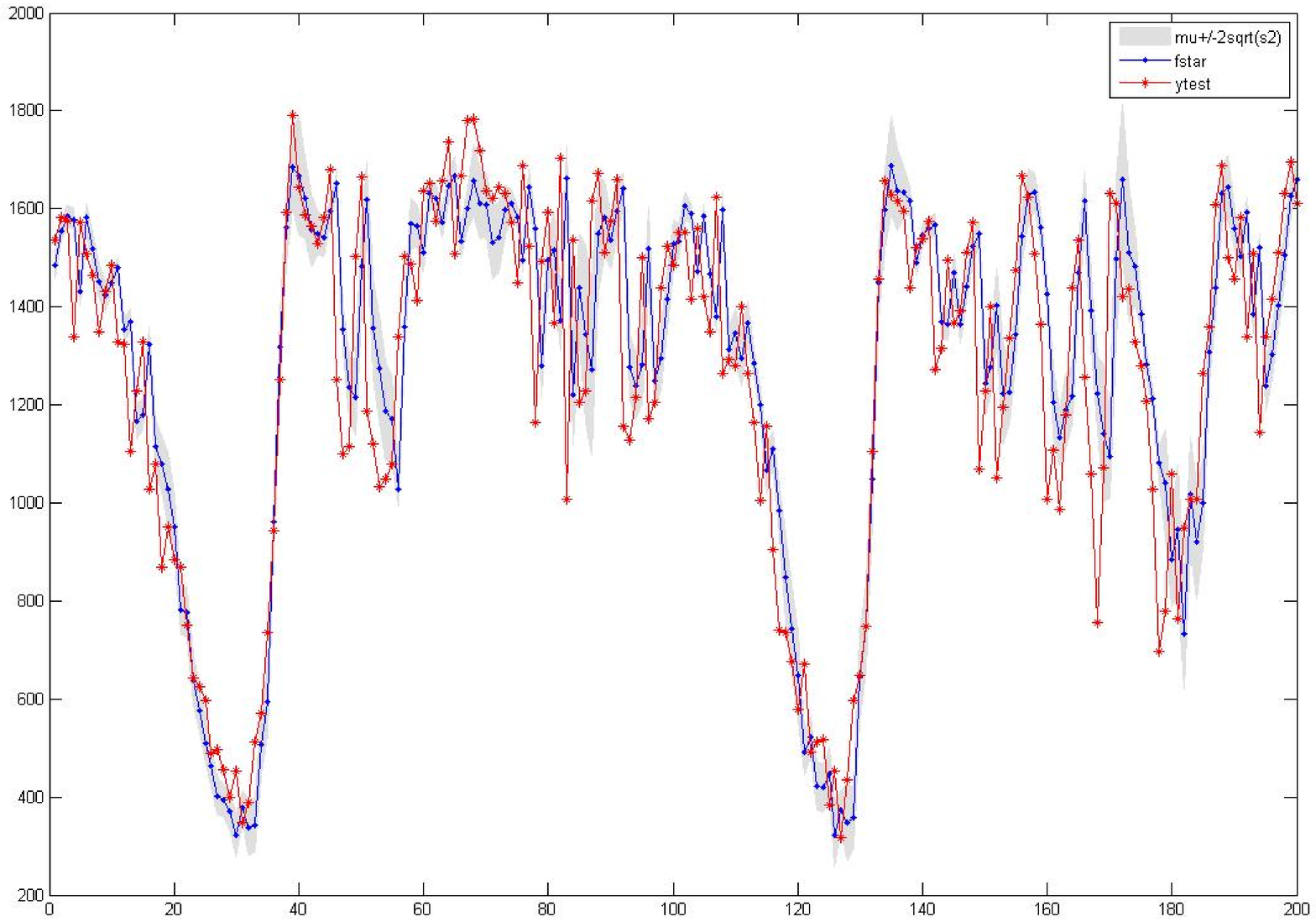}
 \caption{Prediction of link Kd using GPR.}
 \label{Figure 6}
\end{figure}

Reducing the noise variance in Fig.~6, we can get a new prediction
figure of link Kd shown as Fig.~7. From Fig.~7, we can see that the
fluctuating range gets larger when the noise variance is reduced.
The shaded area in Fig.~7 can even contain all the actual targets.
This is the potential capability of GPR. GPR can get more precise predictions by adjusting more appropriate parameters.
Therefore, in the area that it needs to predict only the output scope rather than precise values, GPR is the approach well worth considering.

\begin{figure}[thb]
\centering
\includegraphics[scale=0.35]{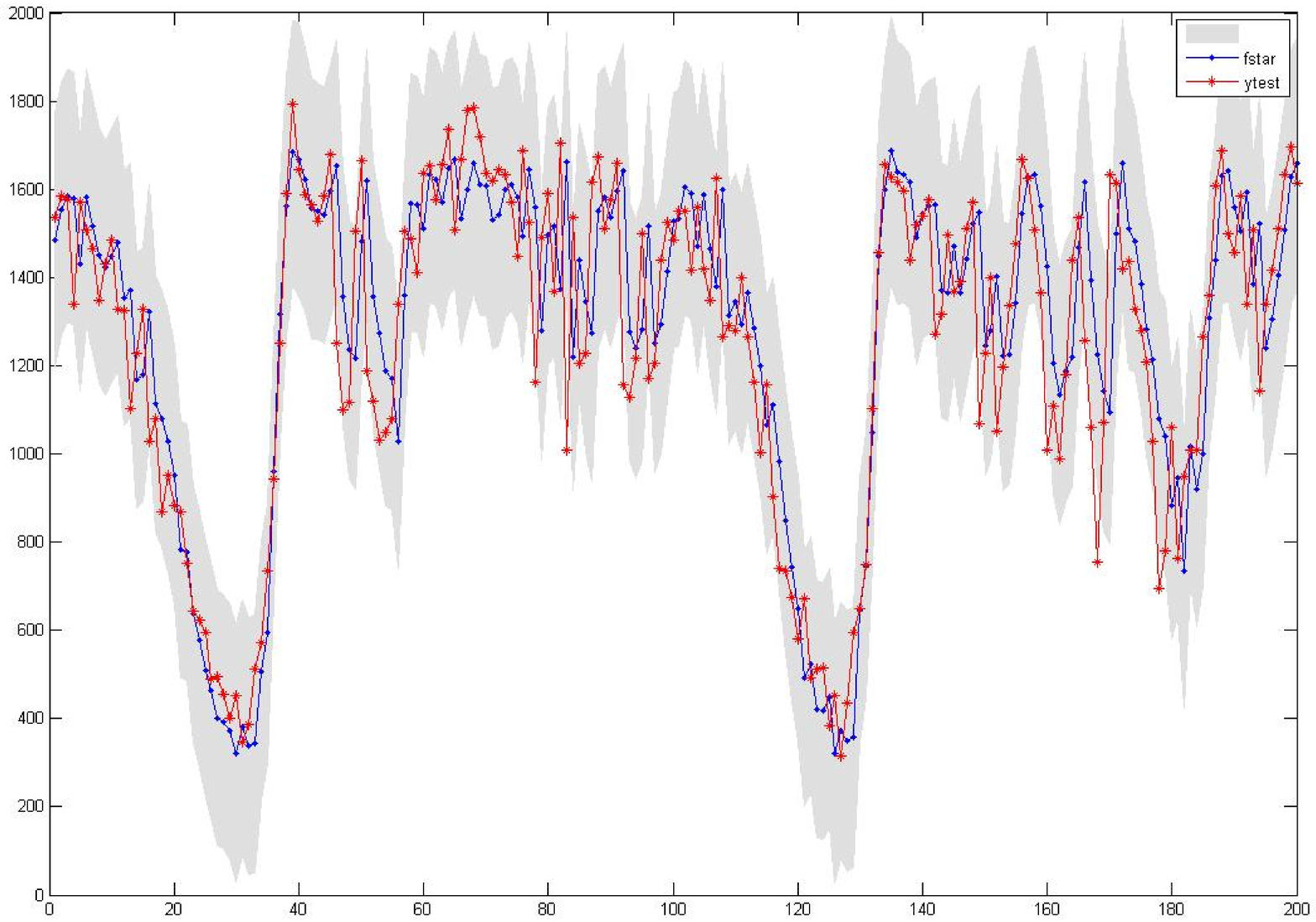}
 \caption{Prediction of link Kd using GPR with reduced noise variance.}
 \label{Figure 7}
\end{figure}

\section*{Conclusions}
Due to the disadvantage of the traditional single-link traffic flow forecasting model,
we propose the multi-link model that predicts traffic flows using historical data from all the adjacent links.
By combining the single-link model and multi-link model with single-task learning and multi-task learning, we propose
four basic traffic flow forecasting approaches, SSTL, SMTL, MSTL and MMTL. Graphical Lasso(GL) is an effective approach in extracting
the relevant information of variables of complex problems by building the sparse graphical model.
We make use of GL to extract the most informative historical flows from all the links in the whole transportation system,
and then construct a BP neural network with the extracted data to predict traffic flows. We refer to the approach combining GL with NN as
GL\_NN. GL\_NN is aslo a multi-link traffic flow forecasting approach, but it is more efficient than MMTL. The test of GL\_NN on real-world traffic
flow forecasting shows competitive results. In addition, we apply GPR to traffic flow forecasting and discuss its potential.
Competitive experimental results reveal the superiority of GL$\_$NN to other proposed approaches.
Moreover, the results further verify that multi-link approaches outperform
single-link approaches and multi-task learning approaches outperform single-task learning approaches in traffic flow forecasting.

In the future, three interesting aspects can be considered. Firstly,
the potential of the multi-link model in traffic flow forecasting
should be further studied. Secondly, GL can be combined with other
approaches not just with NNs. Thirdly, considering that in practical
applications a much lower traffic flow estimation of one road can probably attract vehicles from
adjacent roads and thus causes subsequent traffic difficulties, the
traffic flow prediction methods discussed in this paper can be further
enhanced by investigating microcosmic prediction errors and taking
some precautionary actions for lower estimations.

\section*{Acknowledgements}

This work is supported in part by the National Natural Science
Foundation of China under Project 61075005, and the Fundamental
Research Funds for the Central Universities.


\begin{thebibliography}{99}

\bibitem[Abdulhai et al.(1999)]{AbdulhaiNN99}
Abdulhai, B., Porwal, H., and Recher, W., 1999. Short term freeway flow
prediction using genetically-optimized time-delay based neural
networks, in: \emph{Proceedings of the 78th Annual Meeting of the Transportation Research Board}, Washington D.C., USA.

\bibitem[Abdulhai et al.(2002)]{AbdulhaiNN02}
Abdulhai, B., Porwal, H., and Recher, W., 2002. Short-term traffic flow prediction using neuro-genetic algorithms,
\emph{Journal of Intelligent Transportation Systems}, vol.~7, 3--41.

\bibitem[Banerjee et al.(2008)]{Banerjee08}
Banerjee, O., Ghaoui, L., and Aspremont, A., 2008. Model selection
through sparse maximum likelihood estimation, \emph{Journal of Machine
Learning Research}, vol.~9, 485--516.

\bibitem[Chen and Grant(2001)]{ChenSequential01}
Chen, H. and Grant-Muller, S., 2001. Use of sequential learning for
short-term traffic flow forecasting, \emph{Journal of Transportation
Research Record, Part C: Emerging Technologies}, vol.~9(5), 319--336.

\bibitem[Chen and Chen(2007)]{ChenEnsemble07}
Chen, L. and Chen, C., 2007. Ensemble learning approach for
freeway short-term traffic flow prediction, in: \emph{Proceedings of IEEE
International Conference on System of Systems Engineering}, 1--6. 

\bibitem[Caruana(1997)]{Caruana97}
Caruana, R., 1997. Multitask learning, in: \emph{Proceedings of International
Joint Conference on Machine Learning}, vol.~28(1), 41--75.

\bibitem[Davis and Nihan(1991)]{DavisRegression91}
Davis, G. and Nihan, N., 1991. Non-parametric regression and
short-term freeway traffic forecasting, \emph{Journal of Transportation
Engineering}, vol.~177(2), 178--188.

\bibitem[Davis(1990)]{Davis90}
Davis, G., 1990. Adaptive forecasting of freeway traffic
congestion, \emph{Journal of Transportation Research Record}, vol.~1287, 29--33.

\bibitem[Dahl et al.(2008)]{Dahl08}
Dahl, J., Vandenberghe, L., and Roychowdhury, V., 2008. Covariancem
selection for non-chordal graphs via chordal embedding,
\emph{Optimization Methods and Software}, vol.~23(4), 501--520. 

\bibitem[Duda et al.(2001)]{Duda01}
Duda, R., Hart, P., and Stork, D., 2001. \emph{Pattern
Classification}, John Wiley and Sons, New York.

\bibitem[Friedman et al.(2008)]{Friedman08}
Friedman, J., Hastie, T., and Tibshirani, R., 2008. Sparse inverse
covariance estimation with the graphical lasso, \emph{Biostatistics}, vol.~9(3), 432--441.

\bibitem[Gao and Sun(2010)]{Gao2010}
Gao, Y. and Sun, S., 2010. Multi-link traffic flow forecasting using
neural networks, in: \emph{Proceedings of the Sixth International Conference on Natural
Computation (ICNC)}, 398--401. 

\bibitem[Gao et al.(2011)]{Gao2011}
Gao, Y., Sun, S., and Shi, D., 2011. Network-scale traffic modeling and
forecasting with graphical lasso, in: \emph{Proceedings of the Eighth International
Symposium on Neural Networks (ISNN)}, 151--158. 

\bibitem[Hall and Mars(1998)]{HallArtificial98}
Hall, J. and Mars, P., 1998. The limitations of artificial neural
networks for traffic prediction, in: \emph{Proceedings of the Third IEEE Symposium on
Computers and Communications}, 8--12.

\bibitem[Jordan(2004)]{JordanGraphical04}
Jordan, M., 2004. Graphical models, \emph{Statistical Science}, vol.~19(1), 140--155.

\bibitem[Jin and Sun(2008)]{Jin08}
Jin, F. and Sun, S., 2008. Neural network multitask learning for
traffic flow forecasting, in: \emph{Proceedings of International Joint
Conference on Neural Networks (IJCNN)}, 1898--1902.

\bibitem[Lee and Fambro(1999)]{LeeSubsets99}
Lee, S. and Fambro, D., 1999. Application of subsets autoregressive
integrated moving average model for short-term freeway traffic
volume forecasting, \emph{Journal of Transportation Research Record}, vol.~1678, 179--188.

\bibitem[Moorthy and Ratcliffe(1988)]{MoorthyShort88}
Moorthy, C. and Ratcliffe, B., 1988. Short term traffic
forecasting using time series methods, \emph{Journal of Transportation
Planning and Technology}, vol.~12(1), 45--56.

\bibitem[Meinshausen and B{\"u}hlmann(2006)]{MeinshausenHighDimensional06}
Meinshausen, N. and B{\"u}hlmann, P., 2006. High dimensional graphs
and variable selection with the lasso, \emph{The Annals of Statistics},
vol.~34, 1436--1462. 

\bibitem[Okutani and Stephanedes(1984)]{OkutaniDynamic84}
Okutani, I. and Stephanedes, Y., 1984. Dynamic prediction of
traffic volume through kalman filter theory, \emph{Journal of
Transportation Research Record, Part B}, vol.~18(1), 1--11.

\bibitem[Park et al.(1998)]{ParkRadialBasis98}
Park, B., Messer, C., and Urbanik II, T., 1998. Short-term freeway
traffic volume forecasting using radial basis function neural
network, \emph{Journal of Transportation Research Record}, vol.~1651, 39--47.

\bibitem[Rasmussen and Williams(2006)]{RasmussenGaussian06}
Rasmussen, C. and Williams, K., 2006. Gaussian processes for machine
learning, \emph{the MIT press}.

\bibitem[Smith and Demetsky(1997)]{SmithTraffic97}
Smith, B. and Demetsky, M., 1997. Traffic flow forecasting:
comparison of modeling approaches, \emph{Journal of Transportation
Engineering}, vol.~123(4), 261--266.

\bibitem[Smith and Demetsky(1994)]{SmithTrafficApproaches94}
Smith, B. and Demetsky, M., 1994. Short-term traffic flow prediction
approaches. Prediction: neural network approach, \emph{Journal of
Transportation Research Record}, vol.~1453, 98--104.

\bibitem[Sun and Zhang(2007)]{Sun07}
Sun, S. and Zhang, C., 2007. The selective random subspace predictor
for traffic flow forecasting, \emph{IEEE Transactions on Intelligent
Transportation Systems}, vol.~8(2), 367--373.

\bibitem[Ulbricht(1994)]{UlbrichtMulti94}
Ulbricht, C., 1994. Multi-recurrent networks for traffic
forecasting, in: \emph{Proceedings of the Twelfth National Conference on
Artificial Intelligence}, vol.~1, 883--888.

\bibitem[William and Hoel(2003)]{WilliamSeasonal03}
William, B. and Hoel, L., 2003. Modeling and forecasting
vehicular traffic flow as a seasonal ARIMA process, \emph{Journal of
Transportation Engineering}, vol.~129(6), 664--672.

\bibitem[Wang and Xiao(2003)]{WangRadialBasis03}
Wang, X. and Xiao, J., 2003. A radial basis function neural
network approach to traffic flow forecasting, in: \emph{Proceedings of IEEE
Intelligent Transportation Systems}, vol.~1, 614--617.

\bibitem[Yu et al.(2003)]{YuMarkov03}
Yu, G., Hu, J., Zhang, C., Zhuang, L., and Song, J., 2003. Short-term
traffic flow forecasting based on markov chain model, in: \emph{Proceedings
of IEEE Intelligent Vehicles Symposium}, 208--212.


\bibitem[Zhang and Sun(2010)]{ZhangMultiple10}
Zhang, Q. and Sun, S., 2010. Multiple-view multiple-learner active learning, \emph{Pattern Recognition}, vol.~43, 3113--3119.


\end{thebibliography}
\end{document}